\documentclass[preprint,12pt]{elsarticle}
\usepackage[noend]{algpseudocode}
\usepackage{algorithmicx,algorithm}
\usepackage{amsmath}
\usepackage{multirow}

\usepackage{amssymb}
\usepackage{color}

\begin{document}

\begin{frontmatter}



\title{Dynamic ensemble selection based on Deep Neural Network Uncertainty Estimation for Adversarial Robustness}

%
\tnotetext[1]{This document is the results of the research project funded by the National Natural Science Foundation of China Youth Foundation under Grant 62102449.} 

\author[mymainaddress]{Ruoxi Qin}
\ead[url]{18530023930@163.com}

\author[mymainaddress]{Linyuan Wang}
\ead[url]{wanglinyuanwly@163.com}

\author[mysecondaryaddress]{Xuehui Du}
\ead[url]{dxh37139@sina.com}

\author[mymainaddress]{Jian Chen}
\ead[url]{kronhugo@163.com}

\author[mysecondaryaddress]{Xingyuan Chen\corref{mycorrespondingauthor}}
\cortext[mycorrespondingauthor]{Corresponding author}
\ead{chxy302@vip.sina.com}

\author[mymainaddress]{Bin Yan\corref{mycorrespondingauthor}}
\cortext[mycorrespondingauthor]{Corresponding author}
\ead{ybspace@hotmail.com}

%

\affiliation[mymainaddress]{organization={Henan Key Laboratory of Imaging and Intelligent Processing, PLA Strategy Support Force Information Engineering University},
            addressline={Science Road 62},
            city={Zhengzhou},
            postcode={450001},
            state={Henan},
            country={China}}
\affiliation[mysecondaryaddress]{organization={PLA Strategy Support Force Information Engineering University},
	addressline={Science Road 62},
	city={Zhengzhou},
	postcode={450001},
	state={Henan},
	country={China}}
\begin{abstract}
The deep neural network has attained significant efficiency in image recognition. However, it has vulnerable recognition robustness under extensive data uncertainty in practical applications. The uncertainty is attributed to the inevitable ambient noise and, more importantly, the possible adversarial attack. Dynamic methods can effectively improve the defense initiative in the arms race of attack and defense of adversarial examples. Different from the previous dynamic method depend on input or decision, this work explore the dynamic attributes in model level through dynamic ensemble selection technology to further protect the model from white-box attacks and improve the robustness. Specifically, in training phase the Dirichlet distribution is apply as prior of sub-models' predictive distribution, and the diversity constraint in parameter space is introduced under the lightweight sub-models to construct alternative ensembel model spaces. In test phase, the certain sub-models are dynamically selected based on their rank of uncertainty value for the final prediction to ensure the majority accurate principle in ensemble robustness and accuracy. Compared with the previous dynamic method and staic adversarial traning model, the presented approach can achieve significant robustness results without damaging accuracy by combining dynamics and diversity property.
\end{abstract}

%


\end{frontmatter}


\section{Introduction}\label{sec1}
With the growth of artificial intelligence, deep learning technology has been developed by leaps and bounds, and image recognition performance has continuously improved and gradually surpassed human performance \cite{1}. Due to the extensive deployment of the deep neural network in practical applications, scholars have considered the model's robustness under actual uncertain scenarios. A critical research orientation is to deal with this uncertainty to attain robust prediction results. Data uncertainty may be attributed to the uncertain data scenario due to environmental noise and other factors, and human malice perturbation as a more detrimental factor. In 2014, Szegedy et al. defined this malicious perturbation as the adversarial samples that cause model recognition error without interfering with human eye recognition \cite{2}, triggering the researchers on it. The model gradient-based white box attack method is a standard method, developed from the single-step iteration method as a Fast Gradient Sign Method (FGSM) \cite{3} to the multiple-step iteration method as a Basic Iterative Method (BIM) \cite{4}. Considering the possible unavailability of the model gradient in the actual environment, more research studied the adversarial samples' transferability under the alternative model established based on the same dataset. Based on the iteration method, Momentum Iterative Method (MIM) \cite{5} adds momentum term, the Projected Gradient Descent method (PGD) \cite{6} adds projection operation, while the Diverse Inputs Method (DIM) \cite{7} and the Translation invariant Method (TIM) \cite{8} added input transformation to promote the adversarial examples' transferability. Carlini \& Wagner's (C\&W) approach \cite{9} implemented an effective transfer attack by loss function optimization. A significant threat to the security of deep learning model-based image recognition can be caused by the adversarial samples from those different attack methods.

In order to guarantee the security and availability of deep models in practical applications, the adversarial robustness of deep networks should be realized. Adversarial training \cite{6,10,11,12} is the most effective method at the present stage in robustness research, which damages the model's accuracy and limits the research in single and static model architecture. Such research direction forms an arms race with interdependent and mutually gambled offensive and defensive sides. Some scholars added randomness and dynamics attributes to the model to change the defense methods' passive position in this game. Random smoothing method \cite{13,14,15} based on differential privacy idea added randomness noise to the input to make the gradient of the specific image random in the test stage. Scholars established an actively and adaptively defend method by turning input randomness into the model's tunable parameters. Metzen et al. employed adversarial sample detection to introduce the ability to refuse or accept the prediction results to the model \cite{16}. This method still has the risk of attack due to the static property when relying on an additional detection model. The detection of adversarial samples, which only serves as a criterion to refuse the prediction result, cannot significantly improve the network's robustness. In order to improve the dynamic property, more research introduced the image's learnable bias and optimized it according to the prediction entropy or reconstruction loss in the test stage to achieve the robustness prediction \cite{17,18}. These dynamic methods achieve a process similar to adversarial sample restoration with inefficient forward inference processes. In summary, the study of randomness and dynamics in deep learning models prevents the attacker from ascertaining the gradient of specific images and implementing a white-box attack that changes the defense's passive position in the game to a certain extent.

The above input-based dynamic and randomness methods only treat the image or feature as the variable control object so that the attacker can still achieve an effective attack through its static model and control architecture \cite{19}. Thus, a model-based dynamic control method is an important research direction to improve the dynamic defense method. As a widely studied defense method, the model ensemble can transform the input's dynamic control into the sub-model dynamic selection in the parameter space, thus preventing the attacker from detecting the model's gradient and taking advantage of a black-box attack's inefficient transferability to achieve a more active defense. The decision average-based ensemble technology is the most commonly used approach in deep learning models robustness. In this way, the wrong ensemble prediction can be achieved only if sub-models tend to a similar incorrect prediction, which can guarantee robustness \cite{20}. Therefore, sub-model diversity is essential in ensuring ensemble robustness \cite{21,22}. However, the static ensemble method is the commonly used alternative model for black-box transferability \cite{23}. At the same time, the average outputs strategy cannot adapt to the robustness principle that most of the predictions are accurate. Thus, the weighted and stochastic strategies have been presented for the ensemble \cite{15}. However, the study of sub-model weighted and input-based random smoothing cannot change the ensemble parameters' static property, and the ensemble's dynamic property should be further studied for adversarial robustness.

\begin{figure}[H]
\centering
{\includegraphics[width = \textwidth]{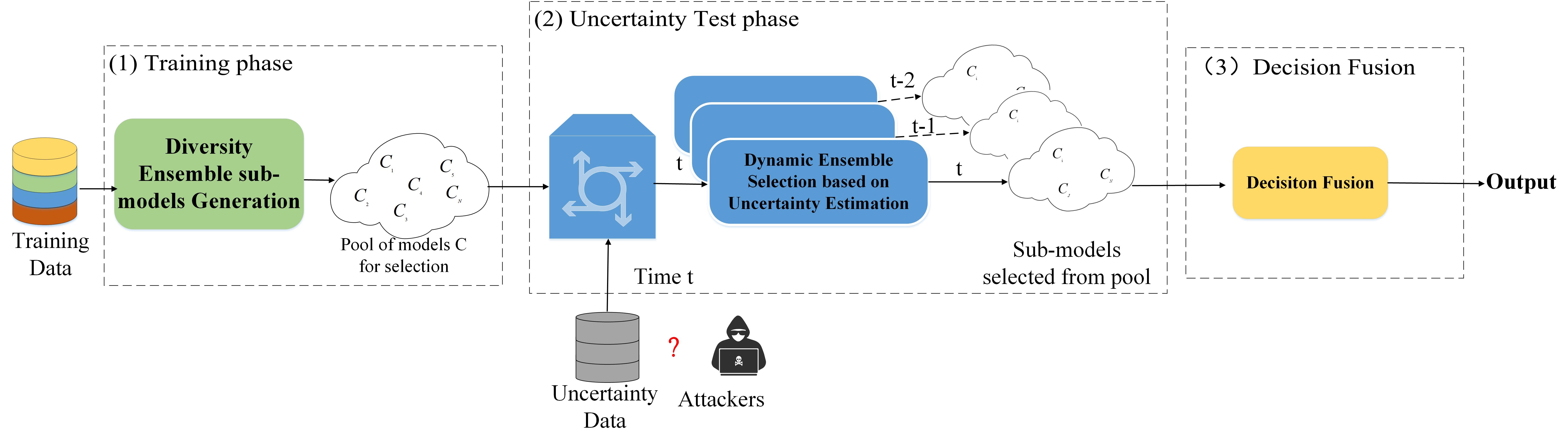}}
\caption{The conceptual flow chart of the proposed ensemble dynamic selection method for adversarial robustness}
\label{fig1}
\end{figure}

This work explores the robustness of the ensemble from the parameter level through sub-model selection based on the uncertainty estimation metric to overcome the robustness limitation caused by single and static ensemble architecture and improve the dynamics to protect model information. As shown in Figure~\ref{fig1}, this work makes each sub-model estimate the uncertainty through Dirichlet prior \cite{24,25,26} in the learning phase and combines the diversity constraints to obtain the sub-model space for selection. In the test phase, the ensemble model will obtain all the sub-mode predictions and estimated uncertainties \cite{27} in forward propagation through the lightweight of the ensemble \cite{28}. The sub-model will be selected after sorting the estimated uncertainties, and its corresponding prediction will be considered the final ensemble result. The presented method takes uncertainty as the model selection criterion based on the diversity sub-models. It does not introduce the optimization problem in the test phase to provide an explicable dynamic policy and testing efficiency, further expanding the robustness brought by the diversity of the ensemble sub-models. In summary, the main contributions of the current work are as the following:
\begin{itemize}
  \item Inspirde by the conventional dynamic ensemble selection technology which widely studied in machine learning, proposed method treats the different convolution models as dynamic attributes to further improve the defense initiative for adversarial samples. 
  \item The ensemble network is fine-tuned using each sub-model output as a prior Dirichlet parameterization over the predictive distribution, providing the uncertainty estimation ability for the individual sub-model. In the test phase, this uncertainty value of each sub-model is used as the dynamic selection policy for robust ensemble prediction.
  \item The pre-trained model is employed to construct the lightweight ensemble network by the low-rank projection. While improving the forward inference efficiency, diversity constraints are introduced among the low-rank projection matrices in parameter space to construct a lightweight, diversified, and easy-to-expand ensemble architecture for dynamic selection. 

\end{itemize}

This paper is arranged as the following. Related works about ensemble architecture, uncertainty estimation, and conventional ensemble selection approach are reviewed in Section~\ref{sec2} to illustrate the main innovation. Then, the details of the proposed ensemble selection for robustness are illustrated in Section~\ref{sec3}. Section~\ref{sec4} presents the experiments and discussion. 

\section{ Related work}\label{sec2}

\textbf{Non-Bayesian-based uncertainty estimation and dynamic decision fusion.} The non-Bayesian uncertainty estimation method utilizes the model's output as variational Dirichlet parameters \cite{24,25,26} and depicts the predicted probability distribution based on its prior conjugate properties of the multinomial distribution. Thus, a single model can estimate the prediction uncertainty through the Evidential theory \cite{27}. This research field has been adopted in the multi-modal and multi-perspective data environment dynamic decision fusion method \cite{36} for enhancing the deep learning model's robustness and precision in the decision space. Most existing research on the robustness of uncertainty estimation methods rely on the possible data uncertainty, and there is no in-depth research on adversarial robustness. The dynamic decision fusion method employed in multi-perspective robustness \cite{36} essentially relies on additional information in data. In contrast, adversarial robustness requires the prediction from the same image, relying more on the information gain brought by the diversity of models. Due to this difference, the prediction results brought by any data are not abandoned in the multi-perspective task but are all fused at the decision level through uncertainty. This work completely abandons the prediction results with high uncertainty according to the majority accurate principle in ensemble robustness. Besides, it makes dynamic attributes at the model level so that such dynamic conditions will be furthest forced to turn the attacker into a black-box condition.

\textbf{Dynamic ensemble selection.} Choosing the appropriate classifier is superior to simply employing the whole multiple classifier systems (MCS) \cite{37}. Dynamic selection (DS) is one of the commonly used MCS methods owing to various experimental works indicating its superiority to static combination approaches. The DS technology estimates the suitable sub-models through the feature space's local area, primarily in the machine learning field. The local region assessment is generally performed by clustering techniques \cite{38,39} and selected by metrics such as accuracy rate \cite{40,41}, ranking \cite{42}, or additional probabilistic models \cite{43,44}. Dynamics here essentially capture uncertainties caused by the consistency and inconsistency between various sub-models. As a result, each single model lacks uncertainty estimation making many inference accesses to verify feature space assessment between test and train samples. Recent studies have proposed a single model's predictive confidence as a new metric for dynamic selection \cite{45}. However, the research did not focus on the deep learning model and the adversarial robustness. Based on the uncertainty estimation of convolutional neural network as the metrics of selection, this paper proposes effective dynamic policy under the deep learning model for adversarial robustness.

\textbf{Diversity and lightweight in the ensemble.} Although a wide range of studies has been performed on robustness and uncertainty estimation of the model ensemble, it may be prohibitive for many applications due to the high cost of computational and memory in the training and testing phases. Dropout approximates the ensemble diversity condition in a single model architecture by randomly deactivating the weight parameters of the fully connected layer during training \cite{29}. Recent studies have employed the Hadamard product through low-rank vectorization projection per sub-model to achieve a lightweight ensemble \cite{28}. From a Bayesian point of view, the Gaussian distribution with learnable parameters is added to model parameters to achieve the lightweight ensemble by sampling the independent model parameters in the forward inference \cite{30,31}. Diversity requirement in light robustness conditions mainly relies on different initializations \cite{32}, different outputs \cite{33}, or stochasticity-injected learning principles \cite{31, 34} for elasticity constrain. In order to simplify the complicated diversity constraint in lightweight conditions, this work transforms the more rigorous diversity constraint based on model parameter repulsive \cite{35} into the repulsive force between low-rank projection matrices.

Inspired by conventional ensemble selection technology, this work further expands robustness through explicit selection metrics in various parameter spaces and improves the current deep learning defense's initiative and reliability.

\section{Approach}\label{sec3}
According to the diversity property, uncertainty estimation is crucial for dynamic ensemble selection robustness. The definition of loss functions is firstly employed to describe the overall construction process of the model. The loss function consists of three parts: the Diversity constraints the differentiation choice space construction of the ensemble; the variational Dirichlet constraints the uncertainty estimation ability of each sub-model; the above two conditions are constrained by uncertainty correction through adversarial and benign samples. In the test phase, In order to achieve the model's robustness, this section further describes the dynamic selection policy by uncertainty estimation of the sub-models in the testing phase of uncertain samples.

\subsection{Model training}
\subsubsection{ Diversity constraints based on low-rank vectorization projection}
The constraint of ensemble diversity is an essential research direction of robustness. D'Angelo et al. \cite{35} first regarded the diversity constraint as a repulsive term during the training phase and derived the relation between diversity and Bayesian property under the parameter space. Equation~\ref{eq1} shows the gradient optimization process while viewing the parameter update of the i-th model with Bayesian posterior probability in the traditional ensemble training.
\begin{equation}\label{eq1}
\begin{array}{l}
\Phi\left(\omega_{i}^{t}\right)=\nabla_{\omega_{i}^{t}} \log p\left(\omega_{i}^{t} \mid D\right) \\
\omega_{i}^{t+1} \leftarrow \omega_{i}^{t}+\varepsilon_{t} \Phi\left(\omega_{i}^{t}\right)
\end{array}
\end{equation}
where $p\left(\omega_{i}^{t} \mid D\right)$ describes the maximum posterior probability estimation of the model parameter $\omega_{i}$ under data $D$. As shown in~\ref{eq2}, the repulsive term \cite{35} employs the RBF kernel function as a distance measure between different values of $\omega_{i}$ to introduce the constraint regularization into the gradient update process.
\begin{equation}\label{eq2}
\begin{array}{l}
\Phi\left(\omega_{i}^{t}\right)=\nabla_{\omega_{i}^{t}} \log p\left(\omega_{i}^{t} \mid D\right)-\mathbb{R}\left(\left\{\nabla_{\omega_{i}^{t}} k\left(\omega_{i}^{t}, \omega_{j}^{t}\right)\right\}_{j=1}^{n}\right) \\
\omega_{i}^{t+1} \leftarrow \omega_{i}^{t}+\varepsilon_{t} \Phi\left(\omega_{i}^{t}\right)
\end{array}
\end{equation}
where $k(\omega_i,\omega_j):\mathbb{R}^d\times \mathbb{R}^d \rightarrow \mathbb{R}$ is the kernel function mapping in the parameter space. Due to the huge parameters of the parameter space in the deep learning model, the Jacobi matrix $J(f_i^t,x)$ obtained from the function space is utilized as the approximate projection of parameters to realize the constraint \cite{35}:
\begin{equation}\label{eq3}
	\Phi\left(\omega_{i}^{t}\right)=\left(\frac{\partial f_{i}^{t}}{\partial \omega_{i}^{t}}\right)^{T}\left[\nabla_{f_{i}^{t}} \log p\left(f_{i}^{t} \mid D\right)-\mathbb{R}\left(\left\{\nabla_{f_{i}^{t}} k\left(J\left(f_{i}^{t}, x\right), J\left(f_{j}^{t}, x\right)\right)\right\}_{j=1}^{n}\right)\right]
\end{equation}

BatchEnsemble \cite{28} constructs the Hadamard product F by the corresponding p-rank vectors $r_i$ and $s_i$ as the projection parameters from the shared weight $\omega$ to the $i$-th sub-model weight $\omega_i$:
\begin{equation}\label{eq4}
\begin{array}{l}
\omega_{i}=\omega \odot F_{i},  { where }\ \mathrm{F}_{i}=r_{i} s_{i}^{T} \\
y=\left(\omega^{T}\left(x \odot r_{i}\right)\right) \odot s_{i}
\end{array}
\end{equation}
Concerning the $\odot$ as element-wise multiplication and kernel function terms K as $k(\omega_i,\omega_j)=exp(-\frac{1}{h}\|(\omega_i-\omega_j) \|^2)$, p-rank vectorization transforms the repulsive terms of the Jacobi matrix into $k(\omega_i,\omega_j)=exp(-\frac{1}{h}\|\omega(F_i-F_j) \|^2)$ through~\ref{eq4}. As shown in~\ref{eq5}, this section divides the repulsive term of F into two terms through r and s, serving as the regularization terms of the model training.

\begin{equation}\label{eq5}
  {Regular}\left(\omega_{i}\right)=\nabla_{\omega_i^{t}} \log p\left(\omega_{i}^{t} \mid D\right)-\mathbb{R}\left(\nabla_{\omega_i^{t}} k\left(r_{i}^{t}, r_{j}^{t}\right)\right)-\mathbb{R}\left(\nabla_{\omega_{i}^t} k\left(s_{i}^{t}, s_{j}^{t}\right)\right)
\end{equation}

When solving $\nabla_{\omega_i}k(:,:)$, the regularization term will force the difference between r or s and force the difference of F, which becomes the diversity constraint of the sub-model in the parameters space. Such constraint eliminates the dependency of the parameter space constraint for diversity on the Jacobian approximation of the function space or the stochasticity-injected learning principle \cite{31,34}. At the same time, the model has a lightweight ensemble and fast-forward inference for training and evaluating the effectiveness.
\subsubsection{Variational Dirichlet and uncertainty estimation}
The output of the traditional Softmax-based single model is regarded as a point estimate of the probability distribution under the Bayesian condition of posterior probability maximization, while the whole posterior probability distribution is not characterized, leading to overconfidence in the prediction and a lack of uncertainty estimation ability. In order to attain a reliable uncertainty estimation for each sub-model, Dirichlet parameterizes a priori distribution over the probability simplex. As shown in~\ref{eq6}, regarding the output $\boldsymbol{\alpha}=[\alpha_i,\ldots,\alpha_N]$ of each sub-model as the Dirichlet distribution parameter, the Dirichlet conjugate priors of the parameter $\boldsymbol{\mu}=[\mu_i,\ldots,\mu_N]$ characterize the multinomial probability distribution $Mult(y^n \mid u^n)$.

\begin{flalign}\label{eq6}
p(\boldsymbol{\mu} \mid x ; \hat{\omega})={Dir}(\boldsymbol{\mu} \mid \boldsymbol{\alpha}), & \quad p(\boldsymbol{y} \mid \boldsymbol{\mu})=Mult(\boldsymbol{y} \mid \boldsymbol{\mu}) \nonumber \\
\boldsymbol{\alpha}=f(x ; \hat{\omega}), \quad {Dir}(\boldsymbol{\mu} \mid \boldsymbol{\alpha})& =\frac{\Gamma\left(\sum_{n=1}^{N} \alpha_{n}\right)}{\prod_{n=1}^{N} \Gamma\left(\alpha_{n}\right)} \prod_{n=1}^{N} \mu_{n}^{\alpha_{n}-1}
\end{flalign}
where $p(\boldsymbol{\mu} \mid x ; \hat{\omega})=Dir(\boldsymbol{\mu} \mid \boldsymbol{\alpha})$ is the variational distribution and $\Gamma(\cdot)$ is the gamma function. In the ensemble model $\omega$ composed of M sub-models' parameters $\omega=\{\omega^m\}_{m=1}^M$, Equation~\ref{eq7} maximizes the evidence lower bound (ELBO) to update the model parameters:
\begin{equation}\label{eq7}
\begin{array}{ll}
\max \limits_{\omega^{m}} L_{m}(x, y)=&\mathrm{E}_{{Dir}\left(\mu^{m} \mid \boldsymbol{\alpha}^{m} ; \omega^{m}\right)}\left[\log p\left(y \mid \boldsymbol{\mu}^{m}\right)\right]\\
&-D_{K L}\left[\operatorname{Dir}\left(\mu^{m} \mid \tilde{\alpha}^{m} ; \omega^{m}\right) \| \operatorname{Dir}\left(\mu^{m} \mid[1, \ldots, 1]\right)\right]
\end{array}
\end{equation}
where the first term of loss is the traditional cross-entropy loss, and the second term sets $\tilde{\alpha}^m=y+(1-y)\odot \alpha^m$ by the classification label y to achieve the Dirichlet prior regularization with the KL Divergence.

According to the variational Dirichlet, the uncertainty can be estimated by differential entropy, which is regarded as a more effective estimation method in various uncertainty scenarios (data and knowledge uncertainties) \cite{24}. The differential entropy is defined in the following equation:
\begin{equation}\label{eq8}
\mathrm{H}^{m}\left[p\left(\mu^{m} \mid x, D\right)\right]=-\int P\left(\mu^{m} \mid x, D\right) \ln \left(P\left(\mu^{m} \mid x, D\right)\right) d \mu^{m}
\end{equation}

The differential entropy is employed to estimate the uncertainty value separately for the single sub-model output. The non-Bayesian uncertainty estimation of different single sub-models can describe different robustness capacities to the same adversarial sample under the diversity condition between sub-models, so the uncertainty value can evaluate the sub-model most likely to classify the adversarial samples correctly. When the sub-models cannot correctly predict all the adversarial samples, the uncertainty estimation can be utilized as the benchmark to eliminate the wrong model interference in the ensemble result.
\subsubsection{Uncertainty correction based on adversarial samples}
Since adversarial samples with stronger concealment significantly differ from noise or corruption samples in their distribution, they construct a particular class of data uncertainty in which its estimation is unreliable \cite{26,46}. In practice, perturbation should be added for additional uncertainty correction \cite{47}. In the model training, the constraint in this section is performed by generating the adversarial samples $x'$ from the benign image $x$ under the BtachEnsemble in the form of adversarial training using 20-step/0.03-eps PGD. The loss is defined as follows:
\begin{equation}\label{eq9}
\begin{array}{ll}
\max \limits_{\omega^{\prime \prime}} R=&\sum_{m=1}^{M} \min \left(\| \mathrm{H}^{m}\left[p\left(\mu^{m} \mid x, \omega^{m}\right)\right]-\mathrm{H}^{m}\left[p\left(\mu^{m} \mid x^{\prime}, \omega^{m}\right)\right] \mid, \gamma\right)\\
&+p\left(y \mid \operatorname{DSC}\left(\alpha^{m} \mid x^{\prime}, \omega^{m}\right)\right)
\end{array}
\end{equation}

The loss includes two main parts. The first part is similar to the margin loss defined by the uncertain correction in \cite{31}. The uncertainty estimation between benign and adversarial samples under the same sub-model is constrained to force their difference variance. $\gamma$ is the threshold for difference variance, set as 8 in this experiment. For the second part of the loss, the Dempster-Shafer Combination (DSC) \cite{36} based on uncertainty estimation is the overall adversarial loss. The robustness loss definition in this way differs from the traditional adversarial training; that is, the ensemble model does not require all sub-models to be robust according to the fusion results of DSC based on uncertainty estimation but requires those with low uncertainty to be adversarial robust. The uncertainty correction force of these loss constraints will be illustrated in detail in the experimental part to describe the difference between adversarial training.
\subsubsection{ Fast fine-tuning based on the pre-trained model}
This work takes the conventional trained baseline model as the deterministic model for low-rank vectorization projection for fast fine-tuning. In the learning process, the ensemble model produces the adversarial samples connected with the benign samples as the input. The input is replicated in channel dimension with the number of integrated models M=10 as a multiple, and all the sub-models are forward inference simultaneously by the grouping convolution \cite{31}. It is worth noting that, unlike the stochasticity-injected learning principle in \cite{31}, this work applies the low-rank projection for all baseline layers and updates it with the same samples. The variational Dirichlet loss of a single sub-model is calculated by~\ref{eq7} for the outputs of benign samples. In contrast, the uncertain estimated calibration loss is calculated by~\ref{eq9} for the output of adversarial samples. Finally, the SGD optimizers $opt_{\omega^m}$ and $opt_{\tilde{\omega}}$, 
\begin{algorithm}[!htp]
	\caption{Training flow of dynamic ensemble selection model} 
	\label{algorithm1}
	\hspace*{0.02in} {\bf Input:}
	Pre-trained model $\tilde{\omega}$, trainging dataset $D\supset (x,y)$\\
	\hspace*{0.02in} {\bf Initialize:}
	Low-rank parameter $\{\omega^m\}_{m=1}^M$, optimizers $opt_{\omega^m}$ and $opt_{\tilde{\omega}}$ \footnotesize\hfill
	\begin{algorithmic}[1]
		\For{$epoch$ in $max\_epoch$} 
		\For{$(x_i,y_i)$ in $D$}
		\State Generate the adversarial example $x'$ from the  sub-models $\{\tilde{\omega}, {\omega}^m\}_{m=1}^M$
		$$x^{'}=x_{t+1}\rightarrow\mathop{\Pi}_{X+s}\left(x_{t}+\epsilon*sign\left(\nabla_{x} J(\omega,x,y))\right)\right)$$
		\State Calculate the Log-likelihood Variational Dirichlet loss $L(x,y)$ via Eq(7)
		\State Obtain the uncertainty correction loss R via Eq(9)
		\State Backward the gradients of total loss $$g_{\tilde{\omega}}=\nabla_{\tilde{\omega}}(L(x, y)+R), \quad g_{\omega^{m}}=\nabla_{\omega^{m}}(L(x, y)+R)$$
		\State Regulate the gradients $Regular(\omega,g_{\omega})$ via Eq(5)
		\State Perform gradient ascent $\tilde{\omega}=\tilde{\omega}+\varepsilon_{1} \cdot {opt}_{\tilde{\omega}}\left(\tilde{\omega}, g_{\tilde{\omega}}\right), \omega^{m}=\omega^{m}+\varepsilon_{2} \cdot o p t_{\omega^{m}}\left(\omega^{m}, g_{\omega^{m}}\right)$
		\EndFor
		\EndFor
		\State \Return $Batchensemble$ $model$ $\{\tilde{\omega},\omega^m\}_{m=1}^M$
	\end{algorithmic}
\end{algorithm}
with learning rates of 0.001 and 0.01, set for pre-trained and low-rank projection parameters in the gradient update stage, respectively. The gradient optimization was regularized with the diversity constraints by~\ref{eq5}. Algorithm~\ref{algorithm1} shows the overall model training process.

\subsection{Dynamic ensemble selection Policy}
In the test phase, the test samples are also replicated in the channel dimension to obtain the output of all sub-models at one time forward inference and different uncertainty estimates for the same sample according to the output of each model through~\ref{eq8}. According to DSC policy \cite{36}, the ensemble model output can be a dynamic fusion in decision level for the output, which is so static and single for the model that it cannot achieve robustness by protecting the model parameters from being detectable for the adaptive attack. Therefore, this work extends the dynamic idea to the model-level selection. Subsequent tests can confirm that this work achieves the dynamics at the model level and obtains better robustness results through model selection.
\begin{figure}[!htp]
	\centering
	{\includegraphics[width = \textwidth]{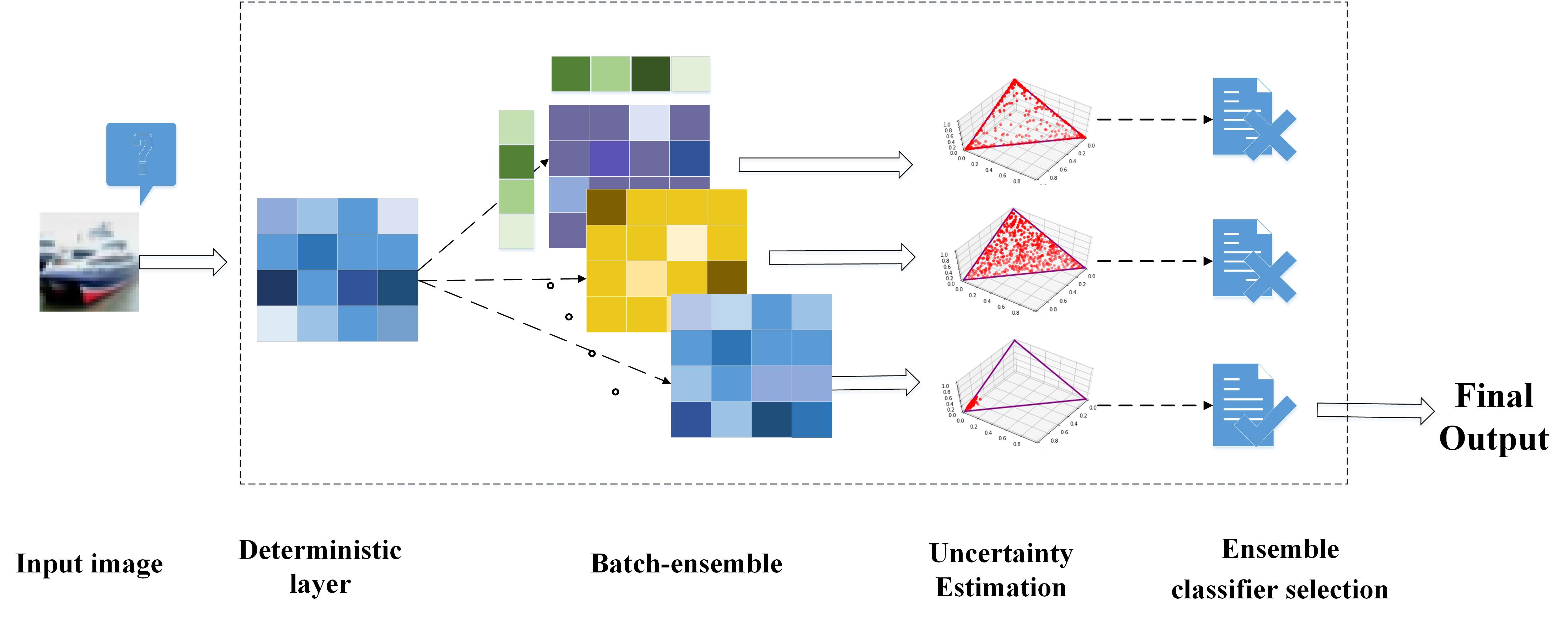}}
	\caption{Ensemble dynamic selection policy based on uncertainty estimation}
	\label{fig2}
\end{figure}

Figure~\ref{fig2} shows the uncertainty estimation-based dynamic ensemble selection method. In the test phase, when each sub-model obtains the uncertainty estimation of the same image through~\ref{eq8}, the output with the minimum uncertainty estimation will be selected as the ensemble model output for the final classification. The model selection is centered on a single image sample, which can differ for each image in an input batch. The selection policy is described as follows:
\begin{equation}\label{eq10}
 { ensemble\_output }=\underset{\mu}{\arg \min }\left\{\mathrm{H}^{m}\left[p\left(\mu^{m} \mid x, \omega^{m}\right)\right]\right\}_{m=1}^{M}
\end{equation}

Such a dynamic selection is based on the greedy policy. The ensemble model takes only one of the optimal estimated model outputs as the ensemble output. The presented method does not set the threshold for selection. On the one hand, the fixed threshold setting is complicated, and the performance of uncertainty estimation-based adversarial sample detection should be improved. On the other hand, a single model selection can maximize the dynamics in the case of a certain number of sub-models. Generally, the model can obtain the uncertainty estimates of different sub-models through on-time forward inference in the test phase, and the sub-model with the minimum uncertainty determines the ensemble output. Given the advantage of BatchEnsemble in efficiency and lifelong learning, model defense can be more active to protect itself form attack in practice. Such dynamic conditions provide the active defense ability for the model when facing attacks and increase the undetectability of model parameters under variable conditions.

\section{Experimental Results}\label{sec4}
The experiment was implemented on the CIFAR-10 dataset \cite{48}, and the Baseline model was WRN-28-10 \cite{49}. For the contrastive model, this section considers different existing defense methods under static (adversarial training), random (random smoothing), and dynamic (closed-loop control and Dent) scenarios using the same baseline architecture. In the training phase, SGD was adopted as the optimizer by contrastive methods, and the batch size was 256. For the presented method, the number of sub-models M was set to 10, the rank of the projection vector p was chosen as 2, the optimizer's learning rate $opt_{\tilde{\omega}}$ for the pre-trained model was chosen as 0.001, the low-rank projection optimizer's learning rate $opt_{{\omega^m}}$ was 0.01, and the iterative training epoch was chosen as 20. All the other contrastive models were trained from scratch, the learning rate was chosen as 0.01, the training epoch was 200, and the learning rate declined by 0.1 at 100 and 150 epochs, respectively. Different experimental settings were trained via the parallel acceleration obtained with 8 RTX3060s, and more detailed parameters were specified in the experimental settings. The experiment reflected the robustness of the proposed method under the transfer-based black-box attack scenario by comparing different selection policies and defense methods. The white-box attack and dynamic experiment were then utilized to demonstrate the superiority of the presented approach and difference to the adversarial training of every single sub-model in the ensemble.
\subsection{Robustness evaluation of different selection policies under the black-box attack}
In the experimental part, the most important selection policy was first discussed through the experiment. This section constructs the experimental model considering that the Baseline is pre-trained without a robustness constraint. For different fusion methods of the ensemble outputs, all outputs can be directly averaged (Average) or fused by DSC at the decision level without any dynamic in model parameters. Different amount of h sub-models are chosen through uncertain rank (uncertain-h) or stochastically (stochastic-h) policies in the uncertain selection policy. This section compares the transfer-based black-box robustness of these different policies. Considering the black-box attacks, the ensemble with different sub-model architectures can achieve better attack transferability as an alternative model \cite{23}. Thus, this section trains the DenseNet121 \cite{50}, Mobilenetv2 \cite{51}, ResNet50 \cite{1}, and Vgg19 \cite{52} under the CIFAR-10 dataset as the alternative model for applying the transfer-based attack. Table ~\ref{tab1} compares the prediction accuracies of different selection policies by benign testing images (None) when applying 50 iterations of Linf adversarial attacks with the maximum perturbations $\varepsilon$= 8/255 and 16/255, respectively. In terms of the methods to counter the attack, this section selects five different and classical transfer attack algorithms (CW, PGD, MIM, TIM, and DIM). The best results are presented in red, and the numbers following the different policies represent the number of employed sub-models.

\begin{table}[!htp]
\center
\resizebox{\linewidth}{!}{\begin{tabular}{lccccccccccccccc}
\hline
\multirow{2}{*}{Policy/Accuracy} &	\multirow{2}{*}{None} &	\multicolumn{2}{c}{CW} &	\multicolumn{2}{c}{PGD} &	\multicolumn{2}{c}{MIM} &	\multicolumn{2}{c}{TIM} &	\multicolumn{2}{c}{DIM}\\
 & & 16&	8&	16&	8&	16&	8&	16&	8&	16&	8\\ \hline
uncertain-1&	90.00&	\textcolor{red}{77.31} &	\textcolor{red}{74.91} &	\textcolor{red}{75.78} &	\textcolor{red}{74.42} &	\textcolor{red}{44.08} &	\textcolor{red}{67.92} &	\textcolor{red}{50.55} &	\textcolor{red}{63.03} &	81.34&	79.01\\
uncertain-2&	\textcolor{red}{90.01} &	74.01&	74.8&	72.57&	74.41&	19.88&	67.75&	23.16&	62.77&	62.37&	79.42\\
uncertain-5&	84.05&	73.83&	69.45&	72.27&	69.40&	19.83&	64.96&	23.10&	58.59&	61.45&	78.35\\
DSC-10&	86.32&	76.56&	71.32&	75.35&	70.70&	43.88&	66.04&	49.01&	60.15&	\textcolor{red}{81.53} &	78.65\\
Average-10&	83.42&	73.75&	68.17&	72.44&	67.71&	21.69&	58.43&	25.12&	55.22&	62.88&	\textcolor{red}{80.21}\\
stochastic-2&	52.58&	22.47&	23.39&	23.11&	24.61&	16.28&	22.57&	18.83&	21.41&	23.86&	24.55\\
stochastic-3&	57.89&	28.64&	28.21&	26.6&	27.94&	18.6&	28.72&	21.73&	26.09&	29.75	&27.88\\
stochastic-5&	67.69&	37.98&	31.16&	40.20&	30.81&	21.58&	28.51&	20.74&	26.94&	33.03&	35.26\\
Proportion&	6.52&	1.60&	1.70&	1.64&	1.68&	1.27&	1.54&	1.26&	1.58&	1.51&	1.77\\ \hline
\end{tabular}}
\caption{Transfer-based black-box attack experiment based on ensemble alternative model under different ensemble selection policies. }
\label{tab1}
\end{table}

Table~\ref{tab1} indicates that the uncertainty estimation-based selection policy has significantly improved the accuracy and robustness compared with stochastic and average methods, demonstrating the ability of uncertainty estimation as selection policy. Although the robustness under DIM is not optimal, it is similar to the optimal defense result. In order to select a sub-model in the uncertainty estimation-based dynamic method, a comparison between different sub-models indicates that selecting a single model through the greedy strategy is the best policy for robustness. Although the accuracy of benign samples does not reach the optimal result, it has a slight difference from the optimal results of selecting the first two models. Compared with the uncertainty-based DSC method to fuse outputs dynamically at the decision level, the dynamic selection at the model level has relatively improved clean accuracy and robustness. According to this adversarial robustness result, although the adversarial training is applied through DSC in section 3.1.3, it cannot provide better robustness than the uncertainty selection-based method. Further, in the last row of Table~\ref{tab1}, the number of accurately predicted sub-models in the ensemble on the black-box adversarial sample is averaged and shown (Proportion), which is not high enough to select randomly for robustness. On the one hand, it illustrates that the introduced adversarial training is more likely to tend to correct the uncertainty estimation for enhancing the robustness of all sub-models. On the other hand, it also indicates that uncertainty estimation contributes to the selection of robust sub-models. A longitudinal comparison of robustness results under different perturbation budgets shows that the higher perturbation budget under MIM decreases its robustness significantly while increasing the number of sub-models to fuse, indicating that blindly increasing the number of sub-models under a powerful attack will likely damage the principle of majority accuracy in the ensemble robustness. An essential property of uncertainty estimation-based dynamic selection is that there is no negative correlation between accuracy and perturbation budget as in previous robustness studies. With the increase of the perturbation budget, the model even has better robust prediction results under the PGD and CW attacks. Although a larger perturbation budget threatens the static model, it may bring a better uncertainty estimation effect for model robustness, which is also one of the advantages of dynamic selection-based uncertainty estimation.

\subsection{Robustness evaluation of different defense methods under the black-box attack}
This section indicates the efficiency of the presented approach by comparing the robustness of different defense methods under the black-box attack. As the most commonly used static defense method, the adversarial training (AT) \cite{10} was applied through the PGD attack with $\varepsilon$=8/255, iteration 10, and step size 1/255. As the most commonly used random defense method, the random smoothing model (RS) \cite{13} was employed by Gaussian noise with Sigma=0.25. Among the dynamic methods, Close-loop Control (CLC) \cite{17} and Defensive entropy minimization (Dent) \cite{18} were selected as typical methods to evaluate the dynamic control robustness. In addition to the dynamic method, the mentioned static and random defense methods were further evaluated through the ensemble of three independently trained models to compare and demonstrate the possible effect of the ensemble on the robustness. In the transfer-based black-box attack, the same method as the previous section was utilized for producing the adversarial samples. Table~\ref{tab2} shows the evaluation results, where the optimal and sub-optimal results are indicated in red and black, respectively.

\begin{table}[H]
\center
\resizebox{\linewidth}{!}{\begin{tabular}{lcccccccccccccccccccc}
\hline
\multirow{2}{*}{Model/Accuracy} &	\multirow{2}{*}{None} &	\multicolumn{2}{c}{CW} &	\multicolumn{2}{c}{PGD} &	\multicolumn{2}{c}{MIM} &	\multicolumn{2}{c}{TIM} &	\multicolumn{2}{c}{DIM}\\
 & & 16&	8&	16&	8&	16&	8&	16&	8&	16&	8\\ \hline
Baseline&	95.82&	2.45&	8.93&	4.40&	11.92	&1.24&	7.48&	4.77&	27.02&	17.6&	54.49\\
AT(0.03)&	73.75&	71.38&	71.38&	71.30&	72.33&	69.93&	71.86&	\textbf{68.44} &	71.26&	69.11&	71.77\\
RS(0.25)&	89.59&	77.07&	80.58&	76.07&	80.70&	30.53&	\textbf{75.98} &	28.48&	70.76&	62.09&	79.58\\
Base ensemble&	95.44&	0.56&	4.11&	1.35&	6.22&	0.35	&3.73&	1.78&	16.3&	14.49&	51.2\\
AT ensemble&	74.84&	72.54&	73.21&	72.30&	73.24&	\textbf{71.27} &	72.91&	\textcolor{red}{69.79}	&\textbf{72.22}	&70.48&	72.99\\
RS ensemble&	85.49&	79.38&	\textcolor{red}{83.50}&	\textbf{78.49} &	\textcolor{red}{82.76} &	31.75&	75.67&	29.67&	71.37&	63.20&	\textbf{82.17} \\
CLC liner& 	91.25&	19.47&	42.86&	18.65&	37.75&	9.82&	16.97&	10.95&	33.35&	18.64&	49.37\\
CLC PMP& 	92.33&	44.91&	57.99&	38.26&	50.45&	15.09&	37.66&	26.89&	49.53&	37.51&	61.66\\
Dent& 	95.62&	44.57&	58.9&	32.71&	47.81&	35.02&	36.44&	16.62&	53.07&	26.96&	69.68\\
Ours&	90.00&	77.31&	74.91&	75.78&	74.42&	44.08&	67.92&	50.55&	63.03&	\textcolor{red}{81.34} &	79.01\\
Ours adv+&	85.11&	\textcolor{red}{83.55} &	\textbf{82.72} &	\textcolor{red}{83.17} &	\textbf{82.66} &	\textcolor{red}{72.67} &	\textcolor{red}{80.18} &	67.96&	\textcolor{red}{78.64} &	\textbf{78.40} &	\textcolor{red}{82.45}\\ \hline
\end{tabular}}
\caption{Transfer-based black-box attack experiment based on ensemble alternative model under different defense methods. }
\label{tab2}
\end{table}

By comparing different defense methods, the random-based smoothing method outperforms the adversarial training in terms of robustness. The accuracy of benign samples mainly confines the adversarial training model's robustness under the transfer-based black-box attack. Ensemble provides a significant improvement for random smoothing but a slight improvement for adversarial training. In contrast, the Baseline ensemble decreases the robustness, indicating the critical role of randomness for ensemble robustness in the case of diversity. The comparison of different dynamic methods indicates that the proposed method achieves the optimal robustness using Baseline as the pre-trained model and rank only second to the random smoothing robustness. Although the accuracy of the presented approach is not optimal under benign samples, it provides the best forward inference and defense efficiency compared with other dynamic methods. Further, by replacing the pre-trained model from the Baseline model with the adversarial training model (adv+), the presented approach attains the optimal transfer-based black-box robustness while ensuring the prediction accuracy of benign samples to a certain extent. Generally, compared with the traditional defense model and its ensemble, the presented method further gives play to the advantages of dynamics through model selection and effectively balances the trade-off between robustness and accuracy.

\subsection{Robustness evaluation under a white-box attack}
A natural question about dynamic model selection is whether it makes sense to choose a model if each sub-model is robust enough. This section compares and analyzes the robustness of the experimental results under the white-box attack to answer this question. Besides, it demonstrates the difference between dynamic selection and robustness brought by adversarial training. In this section, in addition to FGSM, the attack method sets the iteration to 20 and step size to 1/225, while the perturbation budgets are 8/255 and 16/255, respectively. For the ensemble model in the experiment, we assumed that the attacker knows all the ensemble model parameters to take the average output of all the sub-models as the output and generate adversarial samples with Cross-Entropy loss as white-box attack. Table~\ref{tab3} presents the results. The presented method gives the average of correctly predicted values in the ensemble (Proportion), and the optimal robust results are indicated in red.

\begin{table}[H]
\center
\resizebox{\linewidth}{!}{\begin{tabular}{llccccccccccccccccccc}
\hline
\multirow{2}{*}{} & &\multicolumn{2}{c}{FGSM} &	\multicolumn{2}{c}{CW} &	\multicolumn{2}{c}{PGD} &	\multicolumn{2}{c}{MIM} & \multicolumn{2}{c}{TIM} &	\multicolumn{2}{c}{DIM}\\
\multicolumn{2}{c}{} & 16& 8& 16&	8&	16&	8&	16&	8&	16&	8&	16&	8\\ \hline
\multicolumn{2}{c}{Baseline}&	36.62&	51.45&	0.06&	0.28&	0.26&	0.51&	0.70&	1.4&	5.14&	11.43&	23.95&	42.61\\
\multicolumn{2}{c}{AT(0.03)}&	26.86&	46.12&	21.12&	41.1&	14.71&	40.51&	20.04&	41.17&	22.13&	44.84&	38.04&	55.78\\
\multicolumn{2}{c}{RS(0.25)}&	17.45&	27.39&	3.08&	12.44&	1.97&	10.19&	2.72&	11.78&	5.47&	18.43&	29.58&	53.89\\
\multicolumn{2}{c}{Base ensemble}&	28.11&	46.26&	0.65&	1.66&	0.91&	2.41&	1.2&	3.10&	4.94&	13.26&	22.74&	46.14\\
\multicolumn{2}{c}{AT ensemble}&	31.51&	51.28&	29.82&	\textcolor{red}{48.40} &	22.75&	\textcolor{red}{48.09} &	25.20	& \textcolor{red}{48.46} &	29.54&	\textcolor{red}{51.48}	&42.27&	58.80\\
\multicolumn{2}{c}{RS ensemble}&	30.10&	51.02&	7.72&	19.56&	6.60&	19.56&	8.65&	21.79&	13.53&	32.60	&47.78&	68.76\\
\multirow{2}{*}{Ours}&	\footnotesize{Accuracy}&	50.71&	45.07&	3.21&	3.74&	0.45&	0.75	& 2.08&	2.54	&3.16	&6.02	&44.27	&46.15\\
&	\footnotesize{Proportion} &	3.56&	3.42&	0.22	&0.26&	0.03&	0.05&	0.15&	0.18&	0.22&	0.42&	2.97	&3.54\\
Ours&	\footnotesize{Accuracy} &	\textcolor{red}{64.54} &	\textcolor{red}{67.33} &	\textcolor{red}{38.21} &	35.11&	\textcolor{red}{43.32} &	37.36&	\textcolor{red}{34.03} &	37.98&	\textcolor{red}{31.32} &	37.43&	\textcolor{red}{66.53} &	\textcolor{red}{69.17} \\
	adv+ &\footnotesize{Proportion}&	2.58&	2.76&	1.29&	1.25&	1.36&	1.21&	1.12&	1.24 &	1.05 &	1.31 &	2.36	& 2.89\\ \hline
\end{tabular}}
\caption{White-box attack robustness under different defense methods. }
\label{tab3}
\end{table}

The results of traditional defense methods in Table~\ref{tab3} indicate that adversarial training outperforms the random smoothing in terms of robustness under the white-box attack. Under a white-box attack, the random smoothing method can be effectively attacked even though the attacker knows the model gradient exactly but the input. This contradicts the result under the transfer-based black-box attack, indicating that the randomness method for robustness differs from that brought by adversarial training. The smoothing method makes the gradient different from the alternative model under the lack of adversarial training, thus achieving black-box robustness. Gradient protection is the key to robustness in such methods. However, as in our analysis, the input-based randomness attribute has a limited effect on shielding model gradients. Dynamics at the model parameter level can better prevent the model from white-box attacks. The white-box attacker with knowledge of overall parameters is set as an extreme case to analyze the robustness property of the presented approach. The experiment result shows weak robustness under this extreme case, except for the first-order white-box FGSM attack. Such phenomena answer the question raised at the beginning of the current section; that is, the proposed method's robustness property is the same as that of the random smoothing achieved by the gradient difference. Different non-static characteristics only determine an attacker's difficulty in detecting the model gradient rather than the robustness of each sub-model. The competitive white-box attack robustness can be achieved by using adversarial training model as the pre-trained model. Compared with the adversarial training ensemble's robustness results, the presented method provides better robustness under a higher perturbation budget, indicating the critical role of uncertainty estimation in the model's robustness. In summary, the experiments indicate that the robustness of presented method  depends more on the models' dynamics and diversity than a single model. The proposed method employs this characteristic to balance trade-offs between accuracy and robustness.

\subsection{Dynamic analysis}
\begin{figure}[!htp]
	\centering
	{\includegraphics[width = \textwidth]{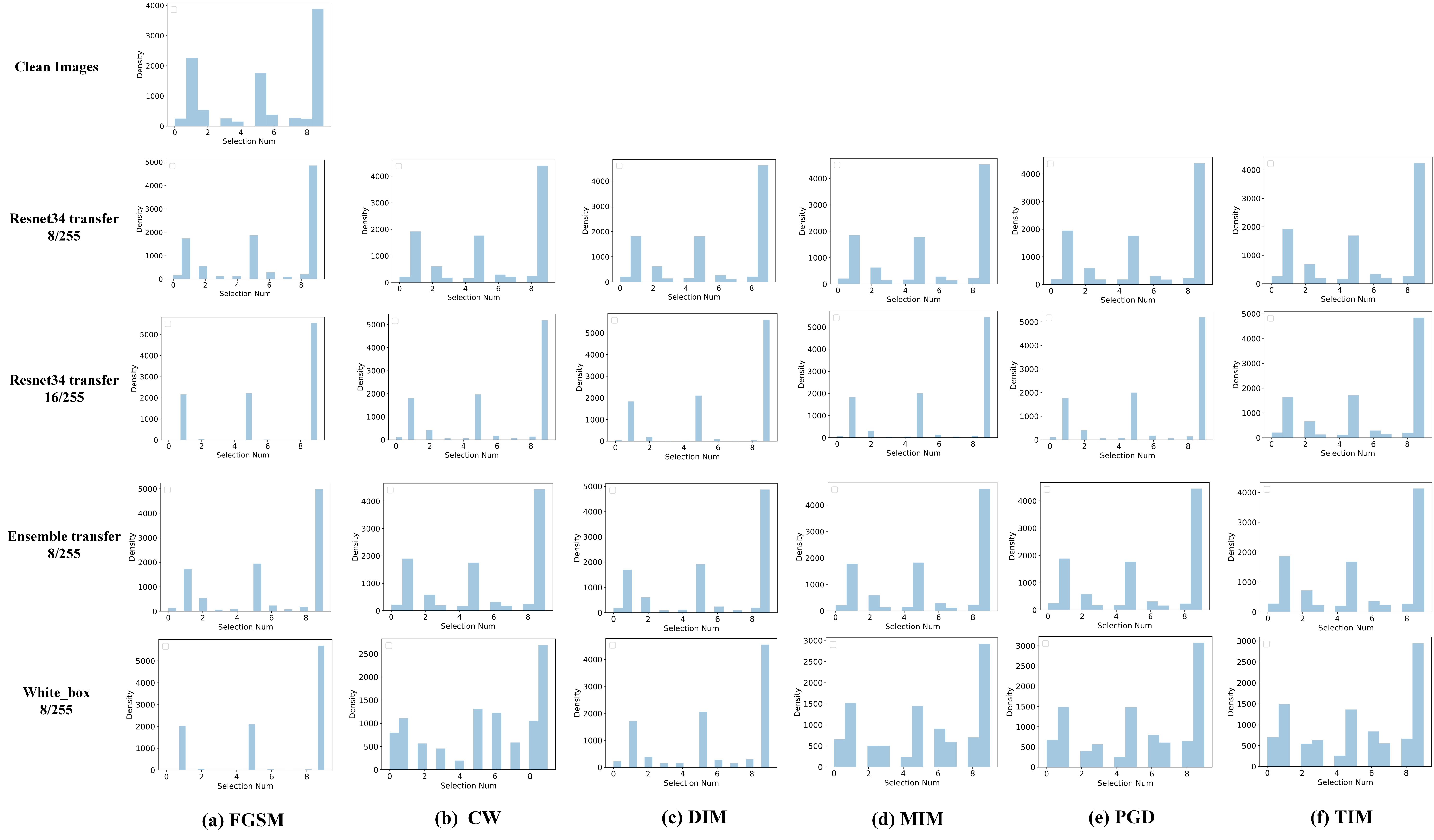}}
	\caption{Histogram of dynamic selection frequency of ensemble sub-models under different attacks.}
	\label{fig3}
\end{figure}

This section shows the histogram statistics of the selection frequency of different sub-models to analyze the model's dynamic selection characteristics. Under 10,000 test images of CIFAR10, the selection frequency of the minimum uncertainty estimate was calculated for different transfer-based alternative models, perturbation budgets, white-box attacks, and benign samples. Figure~\ref{fig3} presents the results.

Although the model selection is not focused on a single model under benign samples, it has specific dynamics in selection. When ResNet-34 is adopted as an alternative model to apply the black-box transfer attack, comparing different attack methods indicates specific differences in model selection under a small perturbation budget, reflecting the model selection's dynamic change according to the change in the attack method. Due to the general increase of the attack strength, as the perturbation budget increases, the model's dynamic no longer varies with the change of attack method. When the ensemble model is adopted as a different alternative model for the transfer attack, the alternative model's transferability becomes stronger. At the same time, it has negligible influence on the model selection under the same perturbation budget. Under a transfer-based black-box attack, uncertainty estimation-based sub-model selection may be more sensitive to the perturbation budget. Under white-box attacks with Cross-Entropy loss function, the MIM, PGD, and TIM attack methods yield results that their selection frequencies differ from the black-box attacks, while their highest selection frequencies are the same as the black-box attacks. Due to the different optimization conditions between CW and other attack methods, its model selection distribution significantly differs from the black-box attacks. Under the white-box attack, the uncertainty estimation-based sub-model selection may be more sensitive to attack conditions optimization. In summary, the proposed model has specific differences and dynamics in the sub-model selection, indicating that the presented approach has the initiative in defense and can prevent the attacker from implementing white-box attacks to a certain extent in practice.

\section{ Conclusion}\label{sec5}
Inspired by the conventional dynamic ensemble selection technology, this work expands the dynamic attribute to the parameter level of deep neural networks and improves the adversarial robustness based on the original intention of dynamic attributes to avoid white-box attacks and increase the defense initiative. A diversified sub-model with uncertainty estimation ability through Dirichlet prior is established for easy expansion and high forward efficiency through a lightweight ensemble of low-rank projection. The ensemble model finally selects the sub-model output with the minimum uncertainty as the output of the whole ensemble in the test phase. The dynamic approach based on uncertainty selection has better robustness under the transfer-based black-box attack than the traditional dynamic approaches. The proposed method combines the adversarial training model with the pre-trained model to obtain more optimal robustness than the traditional random smoothing and static adversarial training models. Since the presented approach relies more on the defensive attributes of diversity and dynamics rather than the robustness of single sub-models, it can overcome the trade-off between precision and robustness brought by adversarial training. In the game of deep learning adversarial attack and defense, the dynamic condition should not take a single model's absolute robustness as the goal in defense but the diversity and dynamic attributes as the goal to increase the budget of white-box attack and end the arms race. The foreseeable future work includes: 1) The threshold of uncertainty estimation can be adjusted to further extend the model's dynamic selection range; 2) The reliability of the model selection policy can be further strengthened by enhancing the sub-models' uncertainty estimation performance to employ them in larger datasets; 3) Given the lifelong learning character of a lightweight ensemble, decision-makers can achieve a dynamic adaptive plan in numerous uncertain environments.





\bibliography{mybibfile}

\end{document}